\documentclass[letterpaper, 10 pt, conference]{ieeeconf}  % Comment this line out if you need a4paper

\usepackage{subfigure}
\usepackage[OT1]{fontenc}
\usepackage{graphicx} % for pdf, bitmapped graphics files
\usepackage{multirow}
\usepackage{algorithm}
\usepackage{algpseudocode}
\usepackage{amsmath}
\usepackage{stfloats}
\usepackage{diagbox}
\usepackage{caption}
\usepackage[colorlinks=true,linkcolor=blue]{hyperref}
\title{\LARGE \bf
SemanticPOSS: A Point Cloud Dataset with Large Quantity of Dynamic Instances
}

\author{\authorblockN
	{Yancheng Pan\authorrefmark{1},
		Biao Gao\authorrefmark{1},
		Jilin Mei\authorrefmark{1},
		Sibo Geng\authorrefmark{1},
		Chengkun Li\authorrefmark{2},
		Huijing Zhao\authorrefmark{1}}
	\authorblockA{\authorrefmark{1}Peking University, Beijing, China}
	\authorblockA{\authorrefmark{2}Beijing Institute of Technology, Beijing, China}
}
\begin{document}

\bibliographystyle{unsrt}

\twocolumn[
{%
	\renewcommand\twocolumn[1][]{#1}%
	\maketitle
	\begin{centering}
		\includegraphics[scale=0.26]{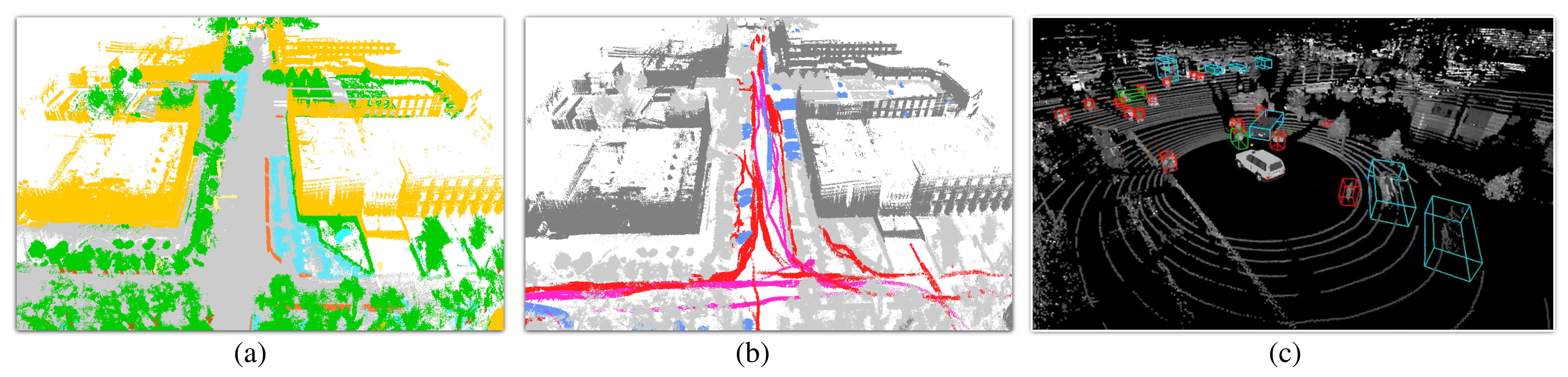}
		\captionof{figure}{A typical scene of SemanticPOSS. (a) LiDAR points of only the static objects are visualized. (b) LiDAR points of the dynamic objects are highlighted. (c) Instances of the dynamic objects in a single LiDAR scan.}
		\label{fig:example}
		\vspace{3mm}
	\end{centering}
}
]

%\maketitle

%%%%%%%%%%%%%%%%%%%%%%%%%%%%%%%%%%%%%%%%%%%%%%%%%%%%%%%%%%%%%%%%%%%%%%%%%%%%%%%%
\begin{abstract}
	
3D semantic segmentation is one of the key tasks for autonomous driving system. Recently, deep learning models for 3D semantic segmentation task have been widely researched, but they usually require large amounts of training data. However, the present datasets for 3D semantic segmentation are lack of point-wise annotation, diversiform scenes and dynamic objects.

In this paper, we propose the SemanticPOSS dataset, which contains 2988 various and complicated LiDAR scans with large quantity of dynamic instances. The data is collected in Peking University and uses the same data format as SemanticKITTI. In addition, we evaluate several typical 3D semantic segmentation models on our SemanticPOSS dataset. Experimental results show that SemanticPOSS can help to improve the prediction accuracy of dynamic objects as people, car in some degree. SemanticPOSS will be published at \url{www.poss.pku.edu.cn}.

\end{abstract}

%%%%%%%%%%%%%%%%%%%%%%%%%%%%%%%%%%%%%%%%%%%%%%%%%%%%%%%%%%%%%%%%%%%%%%%%%%%%%%%%
\section{Introduction}

%\begin{figure*}[b]
%	\centering
%	\includegraphics[scale=0.25]{example.pdf}
%	\caption{A crossing scene of SemanticPOSS. (a) static objects of the crossing in 3D view with dynamic objects hidden. (b) dynamic objects tracking of the crossing. (c) dynamic instances of the crossing in a single LiDAR scan.}
%	\label{fig:example}
%	\vspace{-3mm}
%\end{figure*}

Scene understanding is a vital part of autonomous driving system. In order to distinguish different objects in the environment, autonomous vehicles use sensors such as camera, LiDAR to perceive, and then use a semantic segmentation algorithm to learn how objects distribute in the environment. Compared to camera, LiDAR gives 3D information and it's not easily influenced by various light condition. Therefore, LiDAR has become an essential sensor in most autonomous driving system.

Semantic segmentation for 3D LiDAR point cloud has been widely researched in the past several years. Recently, with the development of deep learning methods, a lot of deep neural network models have been proposed and brought a great progress in 3D semantic segmentation. Some models use 3D point cloud as input directly and give every point a semantic label, such as PointNet\cite{qi2017pointnet}, PointNet++\cite{qi2017pointnet++}, SPLATNet\cite{su2018splatnet}. Others project 3D point cloud to a surface to generate a 2D range image as input, and then give every pixel a semantic label, such as PointSeg\cite{wang2018pointseg}, SqueezeSeg\cite{wu2018squeezeseg}.

However, the generalization performance of deep learning models depends on a large amount and diversity of manually labeled data. For 3D semantic segmentation, we need a large-scale point cloud dataset with point-wise annotation. Although we can find several relevant public datasets such as Semantic3D\cite{hackel2017semantic3d}, SemanticKITTI\cite{behley2019semantickitti}, it's still hard to satisfy the data requirement of deep learning models. In this paper, we try to discuss three problems existing in public datasets for 3D semantic segmentation:

\begin{itemize}
	\item
	Lack point-wise labeled data. Compared to the image datasets such as PASCAL VOC\cite{everingham2015pascal}, ImageNet\cite{deng2009imagenet} and Cityscapes\cite{cordts2016cityscapes}, the scale of point-wise labeled point cloud dataset is limited. As shown in Table.\ref{tab:dataset}, ImageNet contains 14197122 image samples and Cityscapes contains 52425M labeled pixels, much more than any existing 3D dataset. The difficulty of manually annotating 3D point cloud is a nonnegligible reason for the lack of 3D datasets.
	\item
	Lack scene diversity. A single 3D dataset usually have scene selection bias\cite{torralba2011unbiased} in some degree. For example, most 3D datasets only contain scenes of structured urban road or highway environment. The lack of scene diversity will limit the generalization performance of deep learning models. If a model is trained at a dataset with low scene diversity, its performance will drop drastically when testing at new different scenes.
	\item
	Lack dynamic objects. For autonomous driving system, we are extremely concerned about pedestrians, cars in the surrounding. However, most existing 3D datasets contain rich static objects but few dynamic objects, which may make deep learning models detect and recognize the important dynamic instances not as well as other static objects. The quantities of dynamic instances of several datasets are also shown in Table.\ref{tab:dataset}.
\end{itemize}

In order to alleviate the problems mentioned above, we propose the SemanticPOSS dataset, which contains 2988 LiDAR scans with point-wise labeling and 14 classes. Meanwhile, we also give instance-level annotations. The data is collected in Peking University, China. Distinct from typical urban road scenes as most datasets, the campus scenes of SemanticPOSS are various with a large number of pedestrians, riders, cars, and so on. We select an example shown in Figure.\ref{fig:example}.

For convenience, the provided dataset follows the same data format and interface as SemanticKITTI. It almost doesn't need extra data processing for the model training on SemanticKITTI to use SemanticPOSS. We also evaluate some typical 3D semantic segmentation models on our SemanticPOSS dataset.

\begin{table*}[t]
	\centering
	\renewcommand{\arraystretch}{1.5}
	\begin{tabular}{c|cccccc|ccc}
		\hline
		& \multirow{2}{*}{frames}    & \multirow{2}{*}{points}    & \multirow{2}{*}{classes}    & \multirow{2}{*}{type}          & \multirow{2}{*}{annotation}  & \multirow{2}{*}{scene}   & \multicolumn{3}{c}{average instances per frame} \\ \cline{8-10}
		& & & & & & & people & rider & car \\ \hline
		PASCAL VOC\cite{everingham2015pascal} & 9993         & /        & 20  & images   & point-wise  & /   &  / & / & /         \\
		ImageNet\cite{deng2009imagenet} & 14,197,122          & /        & 21841  & images   & bounding box  & /   &  / & / & /         \\
		Cityscapes\cite{cordts2016cityscapes} & 24998         & 52425M         & 30  & images (sequential)   & point-wise / instance  & outdoor   &  6.16 & 0.68 & 9.51         \\
		NYU-Depth V2\cite{silberman2012indoor} & 1449          & 445M         & 894  & RGB-D   & point-wise  & indoor   &  / & / & /         \\
		SUN RGB-D\cite{song2015sun} & 10335          & 3175M         & 800  & RGB-D          & bounding box  & indoor  &  / & / & /          \\
		ScanNet\cite{dai2017scannet} & 2,500,000          & 768000M         & 20  & RGB-D   & point-wise  & indoor   &  / & / & /         \\
		S3DIS\cite{armeni20163d} & 5          & 215M         & 12  & static point clouds       & point-wise    & indoor   &  / & / & /          \\
		Oakland\cite{munoz2009contextual} & 17  & 1.6M  & 44  & static point clouds  & point-wise  & outdoor  & / & / & /   \\
		Paris-lille-3D\cite{roynard2018paris} & 3  & 143M  & 50  & static point clouds  & point-wise  & outdoor &  / & / & /   \\
		Semantic3D\cite{hackel2017semantic3d} & 30  & 4009M  & 8  & static point clouds  & point-wise & outdoor  &  / & / & /   \\
		KITTI\cite{geiger2012we} & 14999  & 1799M  & 8  & sequential point clouds  & bounding box & outdoor  & 0.63 & 0.22 & 4.38 \\
		SemanticKITTI\cite{behley2019semantickitti} & 43552  & 4549M  & 28  & sequential point clouds  & point-wise / instance  & outdoor  & 0.63 & 0.18 & 10.09 \\
		\underline{\textbf{SemanticPOSS}} & 2988  & 216M  & 14  & sequential point clouds  & point-wise / instance & outdoor  & \textbf{8.29} & \textbf{2.57} & \textbf{15.02} \\
		GTA-V\cite{yue2018lidar} & /  & /  & /  & synthetic point clouds  & point-wise & outdoor  & / & / & / \\
		SynthCity\cite{griffiths2019synthcity} & /  & 367.9M  & 9  & synthetic point clouds  & point-wise & outdoor  & / & / & / \\
		\hline
	\end{tabular}
	\caption{The existing main image datasets and 3D datasets.}
	\label{tab:dataset}
\end{table*}

\section{Related Works}

The increasement and improvement of datasets tremendously contribute to the developing of deep learning models. 3D point cloud datasets also help the research of many tasks such as 3D classification, detection and semantic segmentation. Due to the difference of data collection approaches, the existing 3D point cloud datasets can be divided into 4 types: RGB-D, static point clouds, sequential point clouds, synthetic point clouds. Different type of datasets have different density and features of the point clouds.

\subsection{RGB-D Datasets}
RGB-D datasets are collected by RGB-D camera like Kinect. The RGB-D camera not only gets RGB images like common camera, but also perceive the depth for every pixel, to build 3D space information of the scenes. Because it's hard for RGB-D camera to get high-quality data in outdoor scenes, RGB-D datasets only have indoor scenes generally. NYU-Depth V2\cite{silberman2012indoor} contains 1449 RGB-D images from 364 different indoor scenes with pixel-level labels. SUN RGB-D\cite{song2015sun} is similar to NYU-Depth V2, which includes 10335 RGB-D images with 64595 3D bounding box annotations. ScanNet\cite{dai2017scannet} is much more large, which contains 2.5M RGB-D images, 1513 sequences shot in 707 different indoor scenes.

\begin{table*}[b]
	\centering
	\renewcommand{\arraystretch}{1.2}
	\begin{tabular}{cc|ccc}
		\hline
		Sensor    & Maker &  Data   & Specification   & Note \\ \hline
		\multirow{5}{*}{Pandora} & \multirow{5}{*}{Hesai Tech} & \multirow{3}{*}{LiDAR point cloud} & 40-line & \multirow{3}{*}{/} \\
		& & & 0.33 degree vertical resolution & \\
		& & & 200m measurement range & \\ \cline{3-5}	
		& & \multirow{2}{*}{Image} & 4 wide-angle mono cameras & \multirow{2}{*}{Only used for annotation, to be published in future.} \\
		& & & 1 forward-facing color camera & \\ \hline
		\multirow{3}{*}{XW-GI7660 GPS/IMU} & \multirow{3}{*}{StarNeto} & \multirow{3}{*}{Vehicle pose}	& RTK 2cm+1ppm(CEP) & \multirow{3}{*}{Corrected by SLAM\cite{zhao2008slam}.} \\
		& & & Pose accuracy 0.05 degree & \\
		& & & Location accuracy 2cm & \\		
		\hline
	\end{tabular}
	\caption{Sensor configuration.}
	\label{tab:sensor}
\end{table*}

\subsection{Static Point Clouds Datasets}
Static point clouds are some discrete, independent point clouds without dynamic objects, which are collected by laser scanner in two ways. One can put a fixed laser scanner at a specific position to get a quite dense point cloud, or equip a mobile side looking laser scanner and a precise GPS/IMU localization system to generate a dense point cloud covering large areas. This type of datasets usually contains many points but few frames and no dynamic objects. Laser scanners are suitable for both indoor environment and outdoor environment. Stanford Large-Scale 3D Indoor Spaces (S3DIS)\cite{armeni20163d} is composed of 215M points from 5 indoor areas. Oakland\cite{munoz2009contextual} is an outdoor datasets with 1.6M labeled points. Paris-lille-3D\cite{roynard2018paris} consists of 143M points from Paris and Lille. Semantic3D\cite{hackel2017semantic3d} contains 4009M points and different scenes captured in Central Europe.

\subsection{Sequential Point Clouds Datasets}
Sequential point clouds are collected by moving LiDAR. This type of datasets contains a great quantity of points and frames, but the point cloud of a frame is sparse. It's very hard to annotate so many point clouds, so some use 3D bounding boxes to label objects to lighten the workload, such as the KITTI dataset\cite{geiger2012we}. SemanticKITTI\cite{behley2019semantickitti} provides 4549M labeled points in 22 sequences of the KITTI Vision Odometry Benchmark.

\subsection{Synthetic Datasets}
Because of the difficulty of the manual point clouds annotation, synthetic datasets provide an approach to rapidly generate considerable number of point clouds with accurate point-wise labels. GTA-V\cite{yue2018lidar} is a popular video game with a virtual world, whose environment is similar to our real world. One can easily create virtual point clouds in GTA-V by some API. SynthCity\cite{griffiths2019synthcity} is a synthetic point cloud that contains 367.9M points with Gaussian noise. However, there is still a gap between synthetic data and realistic scenes, so synthetic datasets can't completely replace real datasets.

Obviously, it's important to choose proper datasets for different tasks. For autonomous driving system, we are more concerned about sequential point clouds. In addition, we pay more attention to the dynamic instances like pedestrians in the environment. Our SemanticPOSS dataset increases the scale of existing sequential point clouds data with point-wise labels. Besides, the scenes of SemanticPOSS is more complicated with considerable quantity of dynamic instances.

\begin{figure}[b]
	\centering
	\includegraphics[scale=0.27]{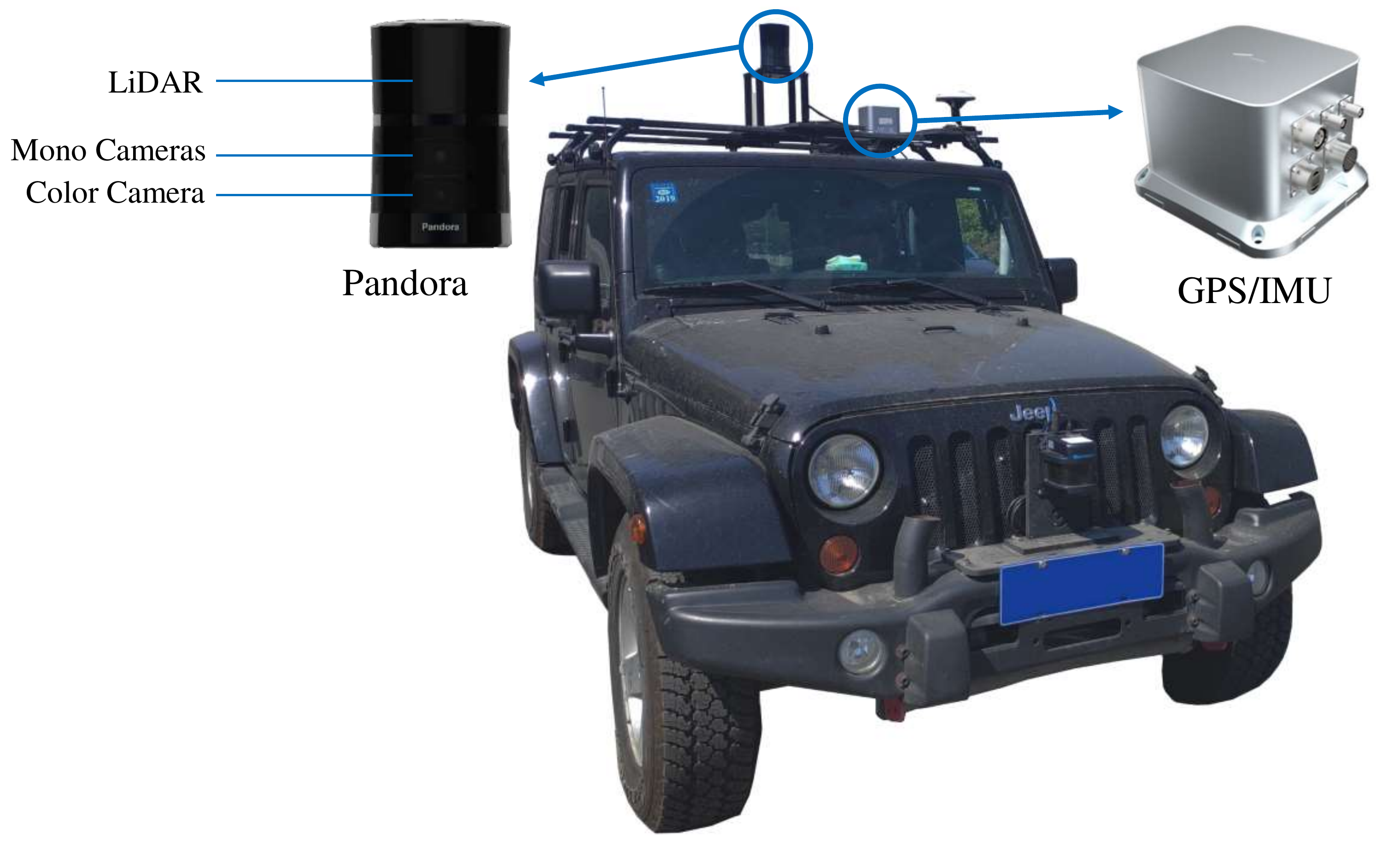}
	\caption{Our data collection vehicle and sensors.}
	\label{fig:sensor}
	\vspace{-3mm}
\end{figure}

\begin{figure*}[t]
	\centering
	\includegraphics[scale=0.3]{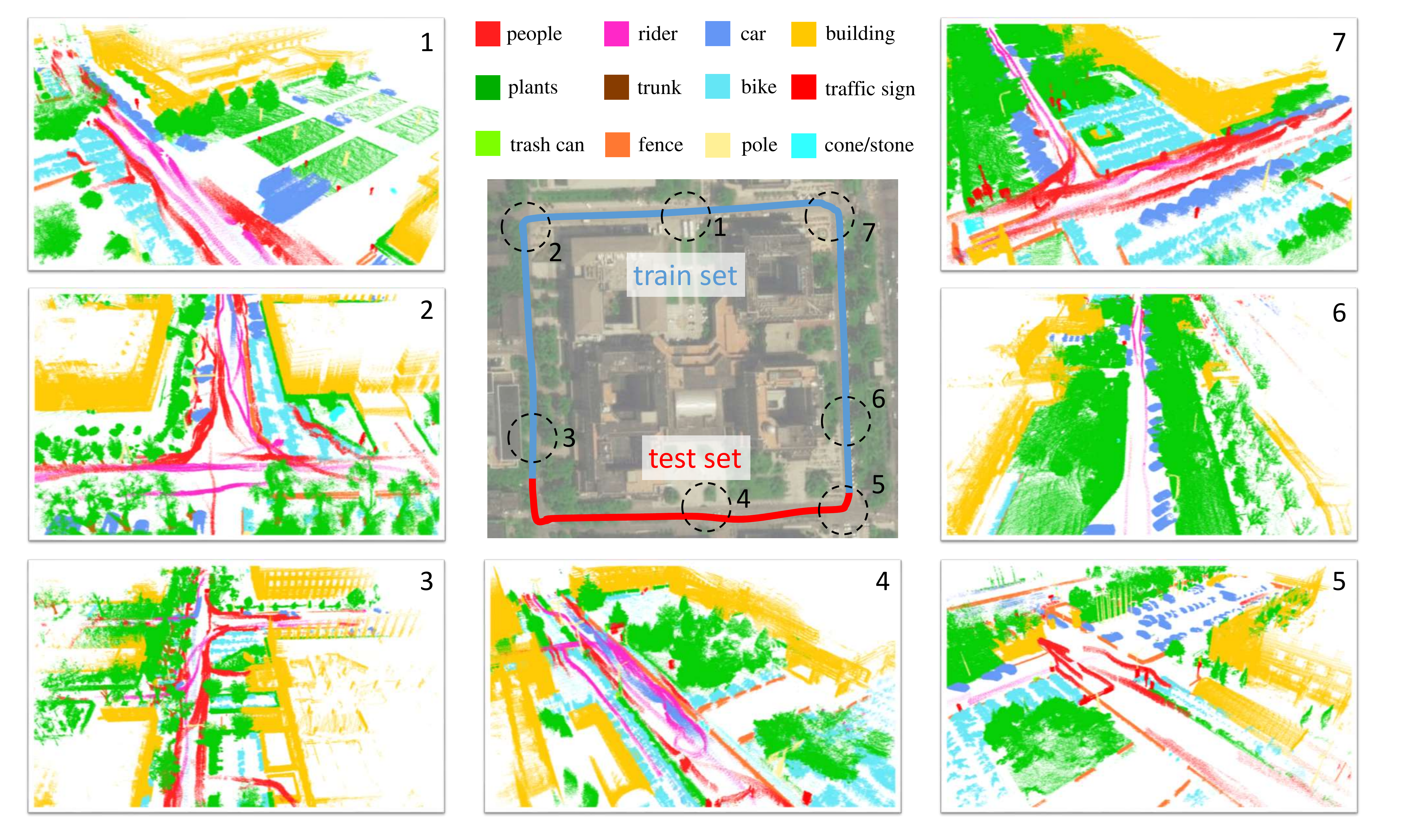}
	\caption{SemanticPOSS dataset. The vehicle's driving route in the campus of PKU, some typical scenes, the train and test sets in experiments.}
	\label{fig:division}
	\vspace{-3mm}
\end{figure*}

\section{Dataset Construction}

\subsection{Sensors and Data Collection}

We use a vehicle equipped with a Pandora\footnote{For more details, please refer to \url{www.hesaitech.com}} sensor module and a GPS/IMU localization system to collect point clouds data. The Pandora integrates cameras, LiDAR and data processing ability into the same module, together with advanced synchronization and calibration features. It consists of a 40-channel LiDAR with 0.33 degree vertical resolution, a forward-facing color camera, 4 wide-angle mono cameras covering 360 degrees around the car. In addition, the GPS/IMU is used for providing localization information to help manual annotation. More details are shown in Figure.\ref{fig:sensor} and Table.\ref{tab:sensor}.

All of the data was collected in Peking University, China. It took about 5 minutes for the data collection vehicle to acquire point clouds, and the vehicle totally traveled about 1.5 kilometers. The vehicle was driven around the teaching buildings and along the way it passed the school gate, main road, parking lot and so on. There were large number of walking or riding students and moving vehicles on the road. Therefore, the scenes are dynamic, various and complicated, especially some road crossing. The route of data collection and some typical scenes are shown in Figure.\ref{fig:division}.

\subsection{Data Annotation}

We annotate every point with semantic label in SemanticPOSS. Moreover, every dynamic object (people, car, rider) has a unique instance label. It's more difficult to label every point in 3D point clouds than label every pixel in 2D images, which is a reason why large-scale 3D dataset is rare. In order to reduce the requirement of manpower and annotation time as much as possible while ensuring label quality, we use the following annotation process shown in Figure.\ref{fig:annotation}:

\begin{figure*}[t]
	\centering
	\includegraphics[scale=0.355]{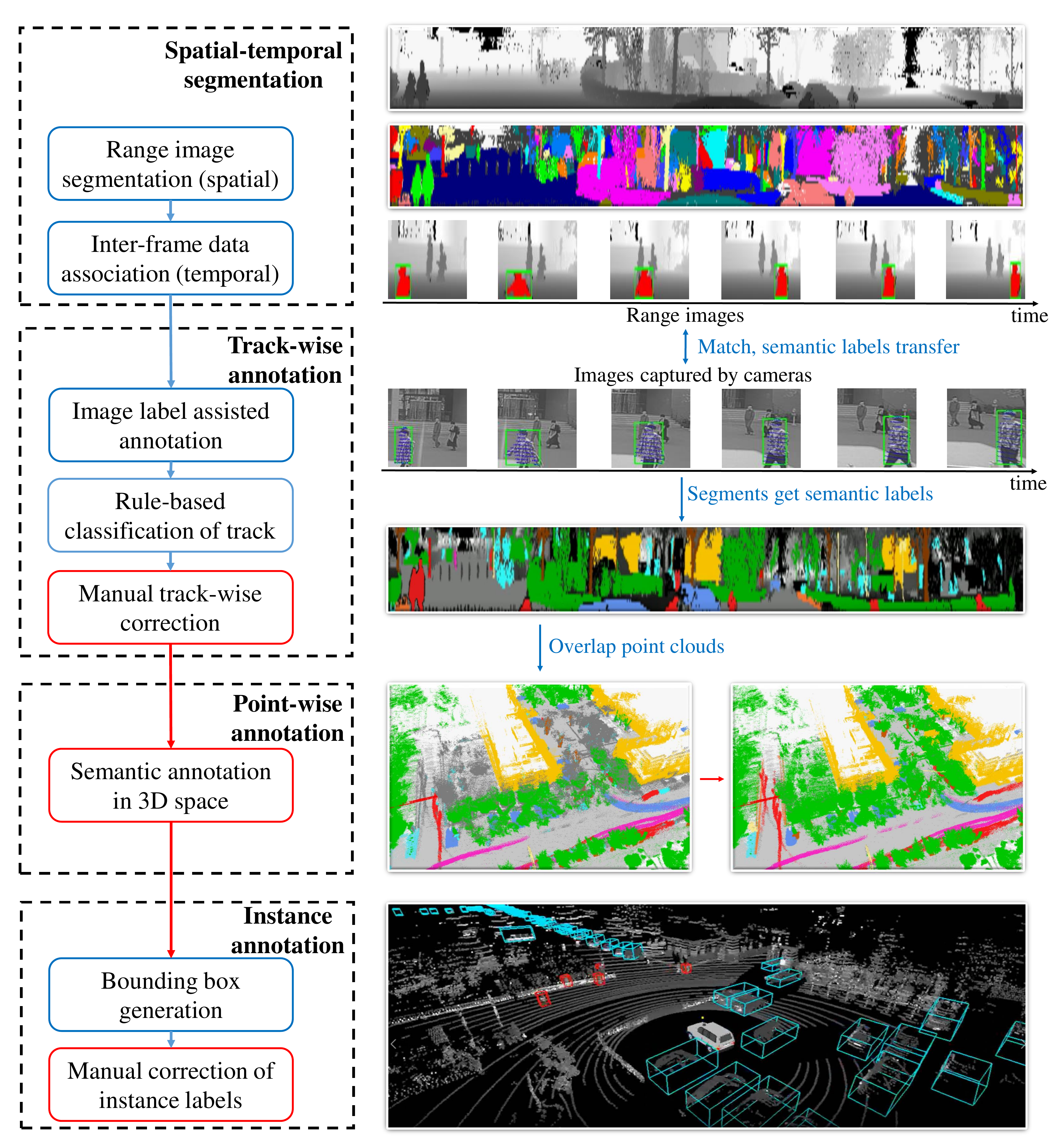}
	\caption{Annotation process. The red boxes and texts mean the work is operated by human, and the blue means the work is operated by computer automatically.}
	\label{fig:annotation}
	\vspace{-4mm}
\end{figure*}

\begin{figure*}[t]
	\centering
	\includegraphics[scale=0.355]{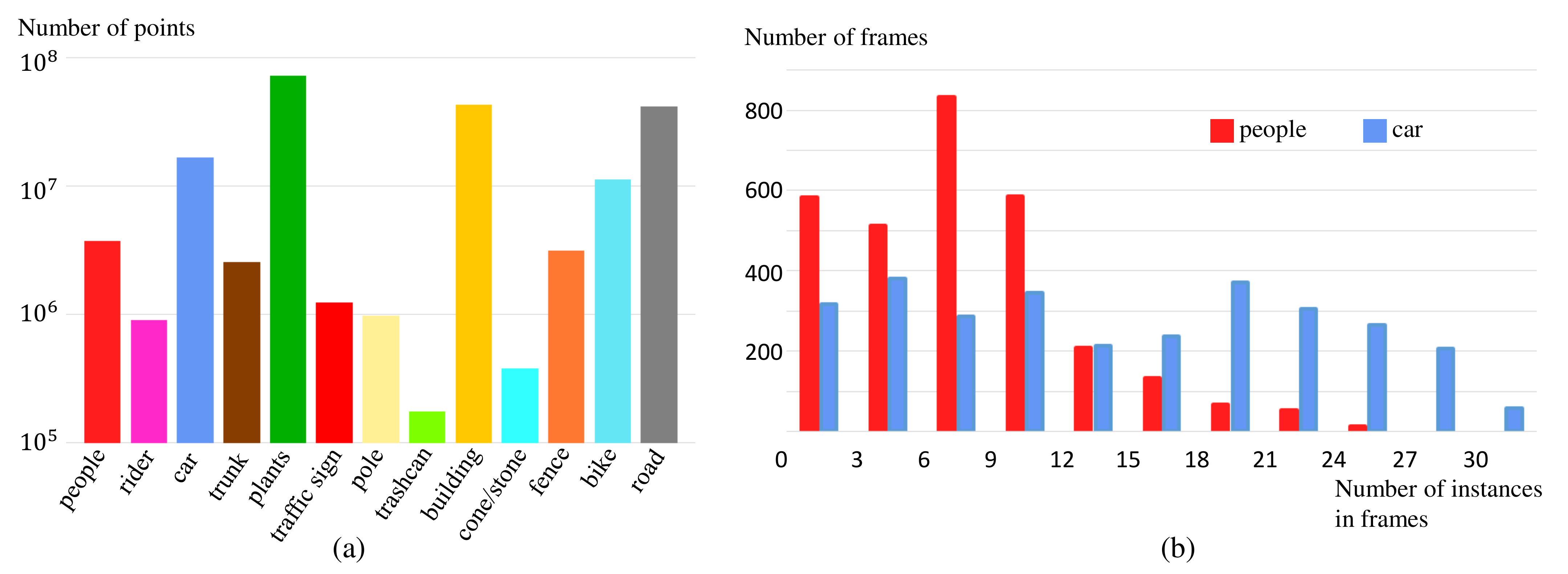}
	\caption{Statistics of SemanticPOSS. (a) The number of points of each class. (b) Histogram of frames containing different number of instances}
	\label{fig:statistics}
	\vspace{-5mm}
\end{figure*}

\subsubsection{Spatial-Temporal Segmentation}
First, we use a segmentation program to divide the point clouds into some segments. It projects the point cloud to a sphere surface to generate a range image. The grey value of a pixel describes the distance between the point and the sensor. It uses the region grow algorithm to divide the range image into several segments, and tracks the same segments in different frames by a simple data association algorithm.

\subsubsection{Track-wise Annotation}
Then, we use a pre-annotation program to give every tracked segment a semantic label by some artificially defined rules. For example, it measures the length, width, height of the segment and finds a proper label to assign from the defined rules. For this track-wise annotation, all points of a tracked segment will get the same semantic label. Meanwhile, it combines information of range images and the matched images captured by cameras to recognize a tracked segment. It uses a pre-trained Mask R-CNN\cite{he2017mask} to annotate people, riders, cars in images and transfers the labels to range images. With the help of the matched images, it will give a relatively good annotation result if the segment is close to the sensor. However, it may give wrong labels or wrong segments, which must be corrected by human. For some unlabeled segments or obviously wrong annotated segments, we manually assign the correct labels to them. And we simply keep the \textit{unlabeled} class to the segment which is hard for us to recognize what it is.

%For convenience and efficiency, segments with same instance label only need one annotating operation, which greatly reduce the required workload. Meanwhile, we show the tracking of each segment with matched image to help annotators recognize the segment easier.

\subsubsection{Point-wise Annotation}
Next, we use the 3D point cloud annotating application provided by SemanticKITTI\cite{behley2019semantickitti} to manually annotate the multiple overlapped point clouds. We overlap several single scan point clouds by calculating the sensor localization of every single scan from the locating information provided by GPS/IMU and calibration. We use SLAM algorithm\cite{zhao2008slam} additionally to acquire a more accurate localization. The annotating application helps us conveniently visualize, operate and annotate the point clouds. Actually, we have already annotated lots of points after track-wise annotation, so we only need to annotate some unlabeled points and misclassified points according to the hint of track-wise annotation result. Because overlapped point clouds are much denser than single scan point cloud, the difficulty of manual annotation decreases remarkably with the help of pre-annotation. However, we need to change our view frequently to annotate points in 3D space. Though only part of the points are required to assign semantic labels, it still takes more than 70 percent of time for us to annotate.

\subsubsection{Instance Annotation}
Finally, we generate the instance labels. After all of the points get right semantic labels, we generate a bounding box for every dynamic instance and redo the data association as the first step. This step can be finished by a computer program automatically and it gives nearly completely correct instance labels without manual intervention. Annotators only need to modify some malposed bounding boxes.

\subsection{Dataset Analysis}

We define 14 classes to label objects. Figure.\ref{fig:statistics}(a) shows the points statistics of every class in SemanticPOSS. There are 2988 point cloud frames with average 8.29 people instances and 15.02 car instances per frame in SemanticPOSS. Figure.\ref{fig:statistics}(b) shows how dynamic instances distribute in frames, where we can find that most frames contain 3-12 people and the number of car varies from frame to frame. The distribution of dynamic instances reflects the variety of the scenes. The large quantity of dynamic objects is the biggest difference between SemanticPOSS and other existing 3D point cloud datasets. For autonomous driving system, it's essential to accurately recognize and detect dynamic objects such as pedestrians and vehicles. Therefore, these complicated and dynamic scenes in SemanticPOSS may help researchers improve the performance and robustness of autonomous driving system in various environment.

\section{Experiments}

\subsection{Evaluation Metrics}

Because our vehicle may pass a position twice or more when collecting data, to ensure the independence between training set and test set, we divide SemanticPOSS according to the location of our vehicle, as shown in Figure.\ref{fig:division}. For annotating convenience, we simply divide SemanticPOSS into 6 parts, 500 frames per part. Data part 3 is defined as test set, and the others are training set.

For 3D semantic segmentation task, deep learning models take point cloud as input and give a semantic prediction to every point. To evaluate the model performance, we use Intersection over Union (IoU) given by

\vspace{-2mm}
\begin{equation}
IoU_{c} = \frac{TP_{c}}{TP_{c}+FP_{c}+FN_{c}}
\end{equation}

where $TP_{c}$,$FP_{c}$,$FN_{c}$ denote the number of true positive, false positive, false negative predictions of the class $c$. Let $N$ be the number of classes used for measure, the mean IoU (mIoU) is defined as the arithmetic mean of IoU, namely

\vspace{-2mm}
\begin{equation}
mIoU = \frac{1}{N}\sum_{c=1}^{N}IoU_{c}
\end{equation}

We design two experiments to analyze the performance of different models and the feature of the dataset itself. The first experiment is training and testing on our SemanticPOSS dataset and comparing the generalization performance of different classes. We just ignore some confusing semantic labels like \textit{unlabeled} when testing. The semantic labels used for evaluation are \textit{people, rider, car, traffic sign, trunk, plants, pole, fence, building, bike, road}. The second experiment is cross-dataset generalization experiment\cite{torralba2011unbiased} between SemanticPOSS and SemanticKITTI, namely training models on SemanticPOSS and testing on SemanticKITTI or contrary. The cross-dataset generalization experiment uses the same semantic labels to evaluate as the first experiment, but some semantic labels defined in SemanticKITTI are ignored or merged to keep classes coincident. Notice that all of the evaluation experiments use single scan point cloud as input, not overlapped point clouds, and weights of all classes are same when training.

\subsection{Baseline Models}

\begin{table*}[t]
	\centering
	\renewcommand{\arraystretch}{1.5}
	\begin{tabular}{c|ccccccccccc|c}
		\hline
		Train and test on SemanticKITTI & people    & rider    & car    & traffic sign   & trunk    & plants  & pole & fence &  building & bike & road  & mIoU\\ \hline
		PointNet++\cite{qi2017pointnet++} & 0.007 & 0.016 & 0.531 & 0.001 & 0.175 & 0.640 & 0.237 & 0.314 & 0.618 & 0.001 & 0.809 & 0.304 \\
		SequeezeSegV2\cite{wu2019squeezesegv2} & 0.154 & 0.393 & 0.784 & 0.201 & 0.322 & 0.726 & 0.192 & 0.404 & 0.697 & 0.122 & 0.882 & 0.470 \\
		\hline
		\hline
		Train and test on SemanticPOSS & people    & rider    & car    & traffic sign   & trunk    & plants  & pole & fence &  building & bike & road  & mIoU\\ \hline
		PointNet++\cite{qi2017pointnet++} & 0.208 & 0.001 & 0.089 & 0.218 & 0.040 & 0.512 & 0.032 & 0.060 & 0.427 & 0.001 & 0.622 & 0.201 \\
		SequeezeSegV2\cite{wu2019squeezesegv2} & 0.184 & 0.112 & 0.349 & 0.110 & 0.158 & 0.563 & 0.045 & 0.255 & 0.470 & 0.324 & 0.713 & 0.298 \\
		\hline
	\end{tabular}
	\caption{Test results of the baseline models in SemanticKITTI and SemanticPOSS.}
	\label{tab:experiment1}
\end{table*}

There are two main deep learning methods for 3D semantic segmentation task distinguished by input form. One uses the raw unordered point cloud as model input and another uses the range image as model input. We choose two typical deep learning models as baseline models, PointNet++\cite{qi2017pointnet++} and SequeezeSegV2\cite{wu2019squeezesegv2}.

PointNet++ is the evolution of PointNet\cite{qi2017pointnet}. PointNet uses raw point cloud as input directly and outputs class of the whole point cloud or semantic labels of every points. It learns point features through max pooling, feature transformations, local and global features combination. PointNet++ ameliorates the problem that PointNet doesn't capture the contextual information of points. PointNet++ applies PointNet on a nested partitioning recursively to generate local features hierarchically, which significantly improve the robustness and generalization performance of the model.

SqueezeSegV2 is derived from SqueezeSeg\cite{wu2018squeezeseg} that uses 2D range image as input to give a point-wise prediction of the point cloud. It achieves a higher prediction accuracy by improving loss function, model structure and robustness to dropout noise of the point cloud. SqueezeSegV2 is based on Convolutional Neural Networks (CNN), Conditional Random Field (CRF) and adds Context Aggregation Module (CAM) to get further improvement. Besides, it boosts the model performance when training on a synthetic dataset and testing on a real dataset.

\subsection{Results and Discussion}

\begin{table}[t]
	\centering
	\renewcommand{\arraystretch}{1.5}
	\begin{tabular}{c|cc|cc}
		& \multicolumn{2}{c|}{mIoU} & \multicolumn{2}{c}{IoU of people} \\ \hline	
		\diagbox{test set}{model} & PN-KITTI    & PN-POSS & PN-KITTI    & PN-POSS   \\ \hline			
		SemanticKITTI & \textbf{0.304} & 0.164 & 0.007 & \textbf{0.064} \\
		SemanticPOSS & 0.127 & \textbf{0.201} & 0.000 & \textbf{0.208}  \\
		\hline
	\end{tabular}
	\caption{Cross-dataset generalization experiment between SemanticPOSS and SemanticKITTI using PointNet++. PointNet++ training on SemanticKITTI is denoted as PN-KITTI, and PointNet++ training on SemanticPOSS is denoted as PN-POSS.}
	\label{tab:experiment2}
\end{table}

\begin{figure*}[t]
	\centering
	\includegraphics[scale=0.38]{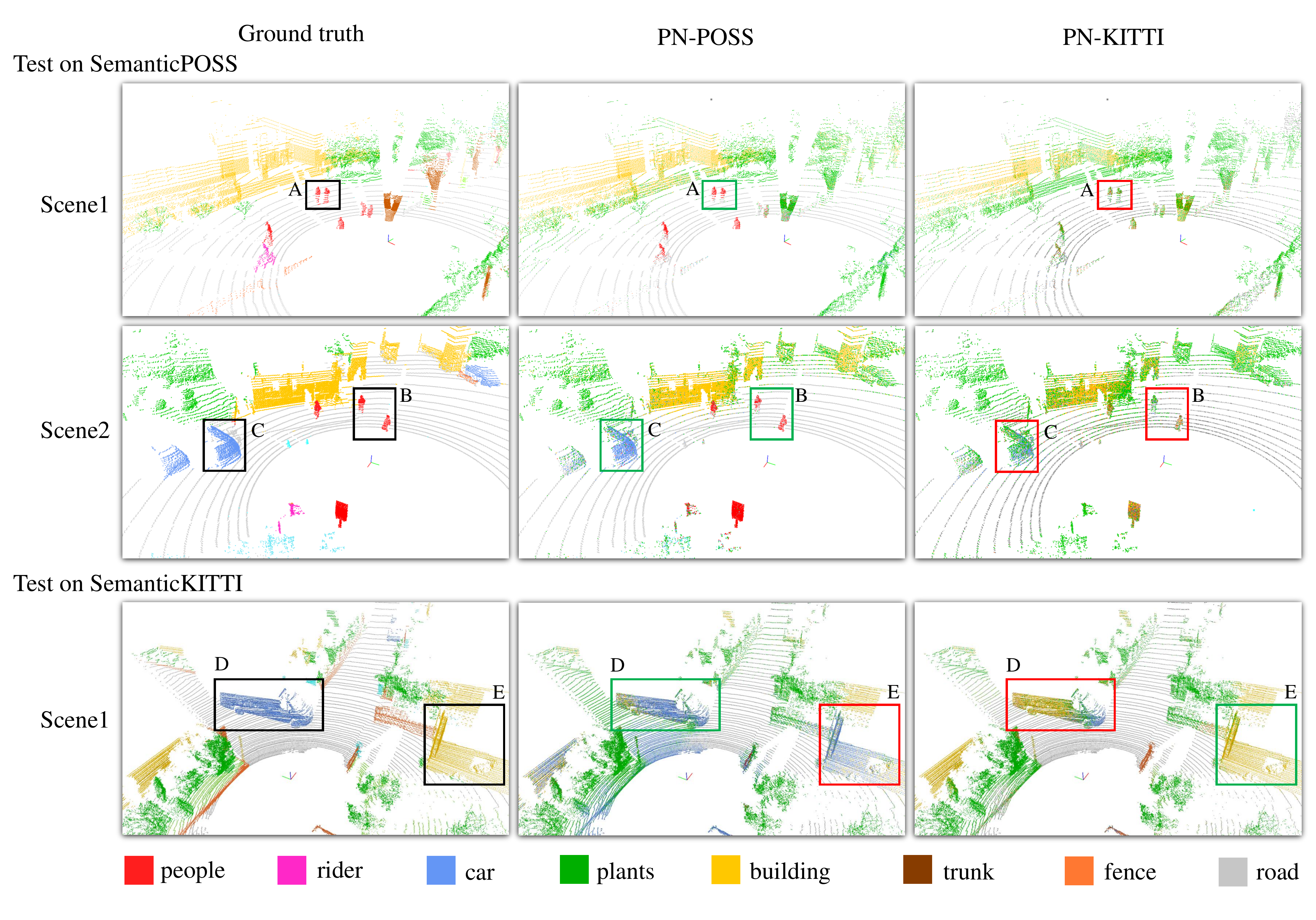}
	\caption{Semantic segmentation results of some scene examples. The abilities to label static objects of PN-KITTI and PN-POSS are not much different. PN-POSS performs better in labeling dynamic objects, as shown in box A, B, C, D. However, it may make mistakes in some static objects, such as box E.}
	\label{fig:result}
	\vspace{-3mm}
\end{figure*}

The results of the first experiment are shown in Table.\ref{tab:experiment1}. For comparison we also show the performance of models training and testing on SemanticKITTI. In order to evaluate whether the variety of scenes have an effect on the performance of deep learning model, we do the second experiment using PointNet++. The results are shown in Table.\ref{tab:experiment2}, where we denote PointNet++ training on SemanticKITTI as PN-KITTI and PointNet++ training on SemanticPOSS as PN-POSS for simplicity. From the experimental results, we have the following findings:

\begin{enumerate}
	\item There is high correlation between IoU and the scale of the corresponding class. For example, the number of road points is much more than the number of people points in datasets, and IoU of road is significantly higher than IoU of people whatever the model we use in Table.\ref{tab:experiment1}. Therefore, sufficiency of data is a necessary condition for deep learning model to learn features of a class.
	
	%\item The model training and testing on SemanticKITTI has higher mIoU than the model training and testing on SemanticPOSS. A primary reason is that the scenes of SemanticPOSS are more complicated. As a consequence, mIoU of PN-KITTI drastically drops when testing on SemanticPOSS in Table.\ref{tab:experiment2}.
	
	\item Both SemanticKITTI and SemanticPOSS reflect a degree of scene selection bias. As a consequence, mIoU of PN-KITTI drastically drops when testing on SemanticPOSS in Table.\ref{tab:experiment2}. Similarly, mIoU of PN-POSS also drops when testing on SemanticKITTI. The scene selection bias can be alleviated if multiple datasets are combines to train models.
	
	\item PN-POSS performs better in labeling dynamic objects. The IoU of people improves when testing on both SemanticKITTI and SemanticPOSS if the model is training on SemanticPOSS. The results show that large number of dynamic instances really improve the abilities to label dynamic objects of deep learning models. Some examples of the model prediction is shown in Figure.\ref{fig:result}.
\end{enumerate}

\section{Conclusion and Future Work}
In this work, we propose a large-scale 3D point cloud dataset with point-wise labels for 3D semantic segmentation task, SemanticPOSS. The dataset contains various and complicated scenes with large quantity of dynamic instances. Through experimental results we find that if training set contains more dynamic instances, deep learning models will get stronger ability to label dynamic objects. Therefore, SemanticPOSS can help deep learning models improve the prediction accuracy of people, car and so on in some degree.

We plan to give further extension to SemanticPOSS in future work. In addition, our Pandora sensor provides quite accurate match between point clouds from LiDAR and images from camera. So we may add images matched with point clouds to SemanticPOSS, to help the research of semantic segmentation methods using multi-sensor fusion.

%\addtolength{\textheight}{-21cm}   % This command serves to balance the column lengths
% on the last page of the document manually. It shortens
% the textheight of the last page by a suitable amount.
% This command does not take effect until the next page
% so it should come on the page before the last. Make
% sure that you do not shorten the textheight too much.

%\bibliography{refs}

\begin{thebibliography}{10}

\bibitem{qi2017pointnet}
Charles~R Qi, Hao Su, Kaichun Mo, and Leonidas~J Guibas.
\newblock Pointnet: Deep learning on point sets for 3d classification and
  segmentation.
\newblock In {\em Proceedings of the IEEE Conference on Computer Vision and
  Pattern Recognition}, pages 652--660, 2017.

\bibitem{qi2017pointnet++}
Charles~Ruizhongtai Qi, Li~Yi, Hao Su, and Leonidas~J Guibas.
\newblock Pointnet++: Deep hierarchical feature learning on point sets in a
  metric space.
\newblock In {\em Advances in neural information processing systems}, pages
  5099--5108, 2017.

\bibitem{su2018splatnet}
Hang Su, Varun Jampani, Deqing Sun, Subhransu Maji, Evangelos Kalogerakis,
  Ming-Hsuan Yang, and Jan Kautz.
\newblock Splatnet: Sparse lattice networks for point cloud processing.
\newblock In {\em Proceedings of the IEEE Conference on Computer Vision and
  Pattern Recognition}, pages 2530--2539, 2018.

\bibitem{wang2018pointseg}
Yuan Wang, Tianyue Shi, Peng Yun, Lei Tai, and Ming Liu.
\newblock Pointseg: Real-time semantic segmentation based on 3d lidar point
  cloud.
\newblock {\em arXiv preprint arXiv:1807.06288}, 2018.

\bibitem{wu2018squeezeseg}
Bichen Wu, Alvin Wan, Xiangyu Yue, and Kurt Keutzer.
\newblock Squeezeseg: Convolutional neural nets with recurrent crf for
  real-time road-object segmentation from 3d lidar point cloud.
\newblock In {\em 2018 IEEE International Conference on Robotics and Automation
  (ICRA)}, pages 1887--1893. IEEE, 2018.

\bibitem{hackel2017semantic3d}
Timo Hackel, Nikolay Savinov, Lubor Ladicky, Jan~D Wegner, Konrad Schindler,
  and Marc Pollefeys.
\newblock Semantic3d. net: A new large-scale point cloud classification
  benchmark.
\newblock {\em arXiv preprint arXiv:1704.03847}, 2017.

\bibitem{behley2019semantickitti}
Jens Behley, Martin Garbade, Andres Milioto, Jan Quenzel, Sven Behnke, Cyrill
  Stachniss, and Jurgen Gall.
\newblock Semantickitti: A dataset for semantic scene understanding of lidar
  sequences.
\newblock In {\em Proceedings of the IEEE International Conference on Computer
  Vision}, pages 9297--9307, 2019.

\bibitem{everingham2015pascal}
Mark Everingham, SM~Ali Eslami, Luc Van~Gool, Christopher~KI Williams, John
  Winn, and Andrew Zisserman.
\newblock The pascal visual object classes challenge: A retrospective.
\newblock {\em International journal of computer vision}, 111(1):98--136, 2015.

\bibitem{deng2009imagenet}
Jia Deng, Wei Dong, Richard Socher, Li-Jia Li, Kai Li, and Li~Fei-Fei.
\newblock Imagenet: A large-scale hierarchical image database.
\newblock In {\em 2009 IEEE conference on computer vision and pattern
  recognition}, pages 248--255. IEEE, 2009.

\bibitem{cordts2016cityscapes}
Marius Cordts, Mohamed Omran, Sebastian Ramos, Timo Rehfeld, Markus Enzweiler,
  Rodrigo Benenson, Uwe Franke, Stefan Roth, and Bernt Schiele.
\newblock The cityscapes dataset for semantic urban scene understanding.
\newblock In {\em Proceedings of the IEEE conference on computer vision and
  pattern recognition}, pages 3213--3223, 2016.

\bibitem{torralba2011unbiased}
Antonio Torralba, Alexei~A Efros, et~al.
\newblock Unbiased look at dataset bias.
\newblock In {\em CVPR}, volume~1, page~7. Citeseer, 2011.

\bibitem{silberman2012indoor}
Nathan Silberman, Derek Hoiem, Pushmeet Kohli, and Rob Fergus.
\newblock Indoor segmentation and support inference from rgbd images.
\newblock In {\em European Conference on Computer Vision}, pages 746--760.
  Springer, 2012.

\bibitem{song2015sun}
Shuran Song, Samuel~P Lichtenberg, and Jianxiong Xiao.
\newblock Sun rgb-d: A rgb-d scene understanding benchmark suite.
\newblock In {\em Proceedings of the IEEE conference on computer vision and
  pattern recognition}, pages 567--576, 2015.

\bibitem{dai2017scannet}
Angela Dai, Angel~X Chang, Manolis Savva, Maciej Halber, Thomas Funkhouser, and
  Matthias Nie{\ss}ner.
\newblock Scannet: Richly-annotated 3d reconstructions of indoor scenes.
\newblock In {\em Proceedings of the IEEE Conference on Computer Vision and
  Pattern Recognition}, pages 5828--5839, 2017.

\bibitem{armeni20163d}
Iro Armeni, Ozan Sener, Amir~R Zamir, Helen Jiang, Ioannis Brilakis, Martin
  Fischer, and Silvio Savarese.
\newblock 3d semantic parsing of large-scale indoor spaces.
\newblock In {\em Proceedings of the IEEE Conference on Computer Vision and
  Pattern Recognition}, pages 1534--1543, 2016.

\bibitem{munoz2009contextual}
Daniel Munoz, J~Andrew Bagnell, Nicolas Vandapel, and Martial Hebert.
\newblock Contextual classification with functional max-margin markov networks.
\newblock In {\em 2009 IEEE Conference on Computer Vision and Pattern
  Recognition}, pages 975--982. IEEE, 2009.

\bibitem{roynard2018paris}
Xavier Roynard, Jean-Emmanuel Deschaud, and Fran{\c{c}}ois Goulette.
\newblock Paris-lille-3d: A large and high-quality ground-truth urban point
  cloud dataset for automatic segmentation and classification.
\newblock {\em The International Journal of Robotics Research}, 37(6):545--557,
  2018.

\bibitem{geiger2012we}
Andreas Geiger, Philip Lenz, and Raquel Urtasun.
\newblock Are we ready for autonomous driving? the kitti vision benchmark
  suite.
\newblock In {\em 2012 IEEE Conference on Computer Vision and Pattern
  Recognition}, pages 3354--3361. IEEE, 2012.

\bibitem{yue2018lidar}
Xiangyu Yue, Bichen Wu, Sanjit~A Seshia, Kurt Keutzer, and Alberto~L
  Sangiovanni-Vincentelli.
\newblock A lidar point cloud generator: from a virtual world to autonomous
  driving.
\newblock In {\em Proceedings of the 2018 ACM on International Conference on
  Multimedia Retrieval}, pages 458--464. ACM, 2018.

\bibitem{griffiths2019synthcity}
David Griffiths and Jan Boehm.
\newblock Synthcity: A large scale synthetic point cloud.
\newblock {\em arXiv preprint arXiv:1907.04758}, 2019.

\bibitem{zhao2008slam}
Huijing Zhao, Masaki Chiba, Ryosuke Shibasaki, Xiaowei Shao, Jinshi Cui, and
  Hongbin Zha.
\newblock Slam in a dynamic large outdoor environment using a laser scanner.
\newblock In {\em 2008 IEEE International Conference on Robotics and
  Automation}, pages 1455--1462. IEEE, 2008.

\bibitem{he2017mask}
Kaiming He, Georgia Gkioxari, Piotr Doll{\'a}r, and Ross Girshick.
\newblock Mask r-cnn.
\newblock In {\em Proceedings of the IEEE international conference on computer
  vision}, pages 2961--2969, 2017.

\bibitem{wu2019squeezesegv2}
Bichen Wu, Xuanyu Zhou, Sicheng Zhao, Xiangyu Yue, and Kurt Keutzer.
\newblock Squeezesegv2: Improved model structure and unsupervised domain
  adaptation for road-object segmentation from a lidar point cloud.
\newblock In {\em 2019 International Conference on Robotics and Automation
  (ICRA)}, pages 4376--4382. IEEE, 2019.

\end{thebibliography}

 % for arXiv

\end{document}